\pdfoutput=1
\documentclass[10pt,twocolumn,letterpaper]{article}

 \usepackage{cvpr}              
\definecolor{cvprblue}{rgb}{0.21,0.49,0.74}
\usepackage[pagebackref,breaklinks,colorlinks,allcolors=cvprblue]{hyperref}
\usepackage[table]{xcolor} 
\def\paperID{10240} 
\def\confName{CVPR}
\def\confYear{2026}

\title{$A^2$GC: $A$symmetric $A$ggregation with Geometric Constraints for Locally Aggregated Descriptors}

\author{Zhenyu Li$^{1, *}$ and Tianyi Shang$^2$\\
$^1$Shandong Academy of Sciences\\
$^2$Fuzhou University
\and
}

\begin{document}
\maketitle
\begin{abstract}
Visual Place Recognition (VPR) aims to match query images against a database using visual cues. State-of-the-art methods aggregate features from deep backbones to form global descriptors. Optimal transport-based aggregation methods reformulate feature-to-cluster assignment as a transport problem, but the standard Sinkhorn algorithm symmetrically treats source and target marginals, limiting effectiveness when image features and cluster centers exhibit substantially different distributions. We propose an asymmetric aggregation VPR method with geometric constraints for locally aggregated descriptors, called $A^2$GC-VPR. Our method employs row-column normalization averaging with separate marginal calibration, enabling asymmetric matching that adapts to distributional discrepancies in visual place recognition. Geometric constraints are incorporated through learnable coordinate embeddings, computing compatibility scores fused with feature similarities, thereby promoting spatially proximal features to the same cluster and enhancing spatial awareness. Experimental results on MSLS, NordLand, and Pittsburgh datasets demonstrate superior performance, validating the effectiveness of our approach in improving matching accuracy and robustness. Our code is available at \textcolor{pink}{https://github.com/CV4RA/A2GC}.
\end{abstract}    
\section{Introduction}
\label{sec:intro}
The ability to recognize a place from a single query image, despite dramatic changes in appearance caused by different times of day, seasons, or viewing angles, remains one of the most demanding challenges in computer vision and robotics \cite{li2025place}. Success in VPR demands image descriptors that remain invariant to the very transformations that make human visual recognition challenging—viewpoint shifts, lighting variations, weather conditions, and seasonal changes, while maintaining the discriminative power necessary to distinguish between visually similar but geographically distinct locations.\\
\begin{figure}
    \centering
    \includegraphics[width=0.85\linewidth]{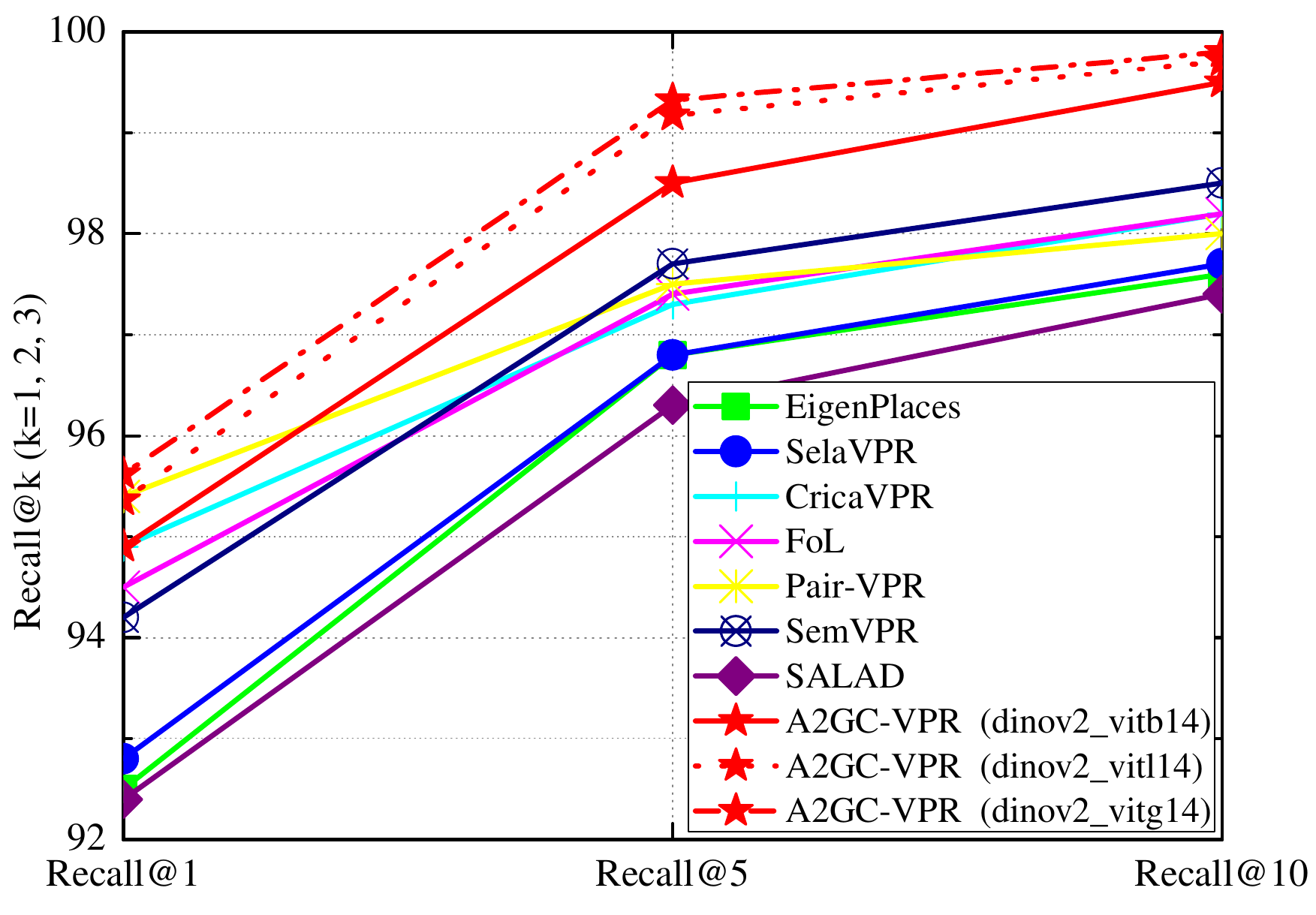}
    \caption{Performance evaluation against state-of-the-art (SOTA) methods based on Pitts30k dataset.}

\end{figure}
Modern VPR pipelines typically follow a two-stage paradigm: first, powerful deep learning backbones extract rich local features from images, then aggregation mechanisms compress these features into compact global descriptors suitable for efficient retrieval \cite{arandjelovic2016netvlad}, \cite{hausler2021patch}. The aggregation stage has emerged as a critical bottleneck, where the choice of how to combine local features profoundly impacts recognition performance. Recent work has explored optimal transport theory as a principled framework for feature aggregation, reframing the assignment of local features to learned cluster centers as a transportation problem \cite{izquierdo2024optimal}. This perspective offers elegant theoretical foundations, but existing implementations rest on a fundamental assumption: that the distributions of image features and cluster centers are symmetric and balanced.\\
\indent This symmetry assumption, however, breaks down in practice. Consider the statistical properties of image features extracted from diverse urban scenes, they may cluster around certain visual patterns, exhibit heavy tails, or show multimodal distributions depending on scene content. The standard Sinkhorn algorithm \cite{cuturi2013sinkhorn}, which enforces symmetric marginal constraints, struggles when these distributions diverge. Additionally, current optimal transport-based methods treat features as independent entities, ignoring the spatial relationships that govern how features are organized within an image. This omission represents a missed opportunity: spatial proximity often correlates with semantic coherence, and leveraging this geometric structure could significantly improve feature assignments.\\
\indent Motivated by these observations, we develop \textit{$A^2$GC-VPR}, a framework that relaxes the symmetry aggregation and incorporates spatial geometry. As shown in Figure 1, our method achieves better performance on the Pitts30k dataset compared to the current SOTA methods, such as EigenPlaces \cite{berton2023eigenplaces}, SelaVPR \cite{lu2024towards}, CricaVPR \cite{lu2024cricavpr}, SALAD \cite{izquierdo2024optimal}, and FoL \cite{wang2025focus}. Our approach introduces two complementary mechanisms: asymmetric aggregation, which adapts to distributional mismatches through row-column normalization averaging and separate marginal calibration, and geometric constraints, which encode spatial relationships via learnable coordinate embeddings and integrate them with feature similarities. Together, these innovations enable more flexible and spatially-aware feature assignments.
\begin{itemize}
    \item We propose an asymmetric aggregation mechanism addressing distributional asymmetry in optimal transport-based feature aggregation. Through row-column normalization averaging with separate marginal calibration, the method accommodates heterogeneous distributions between image features and cluster centers, transcending symmetric optimal transport limitations.
   \item We introduce geometric constraints encoding spatial relationships via learnable coordinate embeddings, computing geometric compatibility scores fused with feature similarities. This enhances assignment locality and coherence by exploiting inherent geometric structure in visual feature representations.
   \item We present $A^2$2GC-VPR, a unified framework integrating asymmetric aggregation with geometric constraints. Experiments on MSLS, NordLand, and Pittsburgh datasets demonstrate state-of-the-art performance with maintained computational efficiency.
\end{itemize}

\section{Related Work}
\label{sec:related work}
\subsection{Feature Aggregation Methods}
Early VPR methods focused on aggregating local features into compact global descriptors. NetVLAD introduced a learnable aggregation layer that generalizes the Vector of Locally Aggregated Descriptors (VLAD) by learning cluster centers and soft-assignment weights. Despite its simplicity, NetVLAD \cite{arandjelovic2016netvlad} established a strong baseline that remains competitive, achieving 90.5\% R@1 on Pitts250k-test. SFRS \cite{ge2020self} improved upon NetVLAD by incorporating spatial feature refinement, achieving 69.2\% R@1 on MSLS-val. CosPlace \cite{berton2022rethinking} addressed the domain gap between training and testing by introducing a cosine similarity-based classifier, demonstrating improved generalization across datasets. The method achieved 92.4\% R@1 on Pitts250k-test and 82.8\% on MSLS-val, establishing a new state-of-the-art at the time. MixVPR \cite{ali2023mixvpr} further advanced aggregation strategies by mixing features across different scales, achieving 94.6\% R@1 on Pitts250k-test and 88.2\% on MSLS-val. SALAD \cite{izquierdo2024optimal} reformulated feature aggregation as an optimal transport problem, considering both feature-to-cluster and cluster-to-feature relations. By introducing a dustbin cluster to discard uninformative features, SALAD achieved 92.2\% R@1 on MSLS-val and 75.0\% R@1 on MSLS-challenge, demonstrating the effectiveness of optimal transport theory in feature aggregation. However, SALAD relies on the standard Sinkhorn algorithm, which assumes symmetric distributions between features and cluster centers, a limitation our work addresses. BoQ \cite{ali2024boq} proposed a bag-of-queries approach, achieving 95.0\% R@1 on Pitts250k-test and 91.4\% on MSLS-val. Despite strong performance, the method lacks explicit handling of distributional asymmetry.\\
\indent \textit{Despite significant advances, existing methods face several limitations. First, aggregation-based methods typically assume symmetric distributions between features and cluster centers, which may not hold in practice. Second, many methods ignore spatial relationships between features, missing opportunities to leverage geometric structure.}
\subsection{Transformer-based Approaches}
The success of transformer architectures in computer vision has led to their adoption in VPR. R2Former \cite{zhu2023r2former} introduced a recursive refinement mechanism using transformers, achieving 93.2\% R@1 on Pitts250k-test and 89.7\% on MSLS-val. The method's recursive refinement process enables iterative improvement of feature representations, demonstrating strong performance on MSLS-challenge with 73.0\% R@1. SelaVPR \cite{lu2024towards} employed self-learning mechanisms with transformer architectures, achieving state-of-the-art results: 95.7\% R@1 on Pitts250k-test, 90.8\% R@1 on MSLS-val (the highest at the time), and 73.5\% R@1 on MSLS-challenge. The method's self-learning approach enables adaptation to diverse visual conditions.  CricaVPR \cite{lu2024cricavpr} introduced cross-image attentive matching mechanisms, achieving 95.6\% R@1 on Pitts250k-test and 99.5\% R@10 (the highest R@10). The method demonstrated strong performance across datasets but requires additional computational overhead for cross-image attention. EigenPlaces \cite{berton2023eigenplaces} leveraged eigenvector-based representations for place recognition, achieving 94.1\% R@1 on Pitts250k-test and 89.1\% on MSLS-val. The method's geometric approach provides interpretable representations. \\
\indent \textit{Unlike transformer-based methods that require architectural modifications, our approach addresses these limitations by introducing asymmetric aggregation that adapts to distributional discrepancies, and geometric constraints that encode spatial relationships, and also integrates seamlessly with existing aggregation frameworks, maintaining computational efficiency while achieving competitive performance.}

\section{Approach}
In this section, we present $A^2$GC-VPR, a unified framework that integrates asymmetric aggregation with geometric constraints for visual place recognition. We first formulate the feature aggregation problem within the optimal transport framework, then introduce our asymmetric aggregation mechanism and geometric constraints, and finally describe the complete pipeline shown in Figure 2.
\begin{figure*}[htbp]
    \centering
    \includegraphics[width=0.95\linewidth]{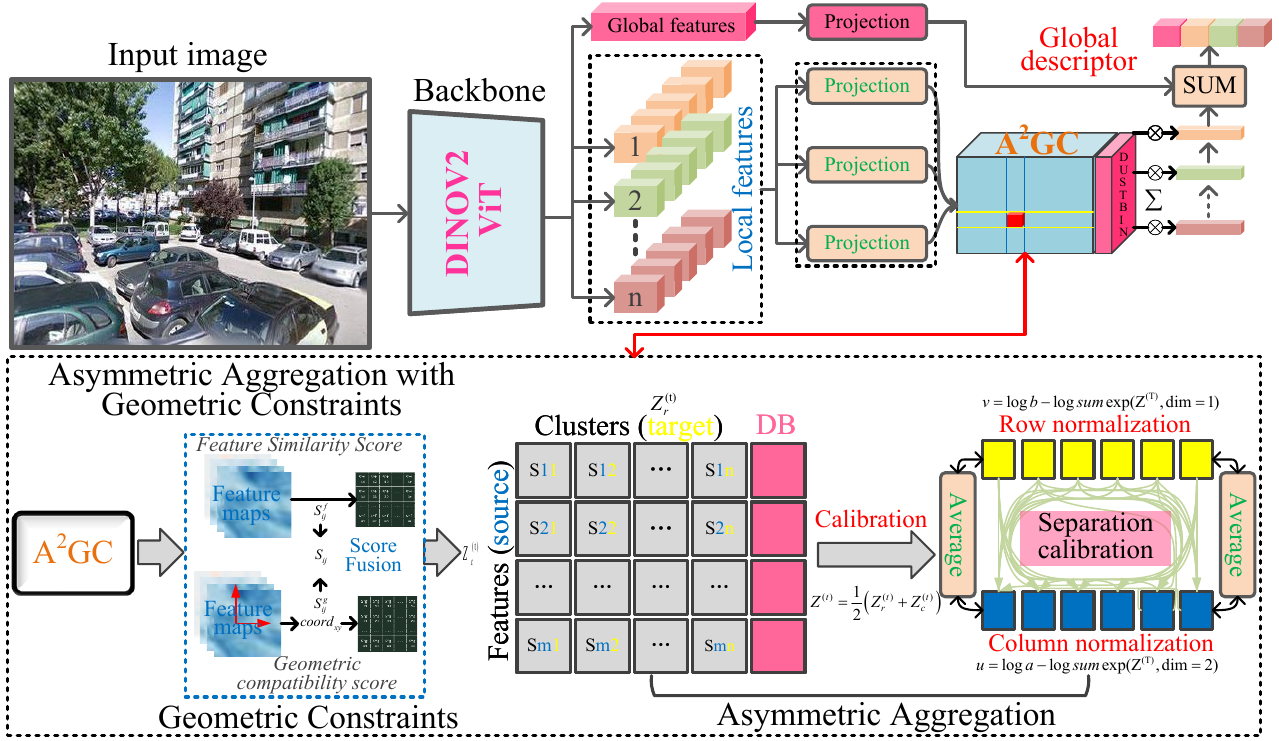}
    \caption{The Illustration of $A^2$GC pipeline. $A^2$GC uses DINOv2 to extract local features and a global token. Geometric constraints are integrated by encoding spatial coordinates as learnable embeddings and computing geometric compatibility scores, which are fused with feature similarities. Also, $A^2$GC employs row-column normalization averaging, separate marginal calibration, and determines the transport matrix that assigns features to clusters. The aggregated cluster descriptors are concatenated with the projected global token and normalized to form the final global descriptor.}
\end{figure*}
\subsection{Problem Formulation}
Given an input image $I$, we extract local features $\mathbf{F} = \{\mathbf{f}_i \in \mathbb{R}^d\}_{i=1}^n$ from a deep backbone network, where $n = H \times W$ is the number of spatial locations and $d$ is the feature dimension. The goal is to aggregate these local features into a compact global descriptor $\mathbf{q} \in \mathbb{R}^{m \cdot l + g}$ that captures discriminative information for place recognition.\\
\indent Following SALAD \cite{izquierdo2024optimal}, we reformulate feature aggregation as an optimal transport problem. Let $\mathbf{C} = \{\mathbf{c}_j \in \mathbb{R}^d\}_{j=1}^m$ denote a set of $m$ learnable cluster centers, and $\mathbf{P} \in \mathbb{R}^{m \times n}$ be a transport matrix that assigns each local feature $\mathbf{f}_i$ to cluster centers. The standard optimal transport formulation seeks to minimize:
\begin{equation}
\min_{\mathbf{P}} \sum_{i,j} \mathbf{P}_{ij} \mathbf{C}_{ij} - \epsilon H(\mathbf{P})
\end{equation}
subject to marginal constraints:
\begin{equation}
\sum_{j} \mathbf{P}_{ij} = \mathbf{a}_i, \quad \sum_{i} \mathbf{P}_{ij} = \mathbf{b}_j
\end{equation}
where $\mathbf{C}_{ij}$ represents the cost of transporting feature $\mathbf{f}_i$ to cluster $\mathbf{c}_j$, $H(\mathbf{P})$ is the entropy regularization term, and $\mathbf{a}$, $\mathbf{b}$ are the source and target marginals, respectively.\\
\indent However, the standard \textit{Sinkhorn} algorithm assumes symmetric marginal constraints, which may not hold when image features and cluster centers exhibit different distributions. This limitation motivates our asymmetric aggregation mechanism.
\subsection{Asymmetric Aggregation Mechanism}
To address the distributional asymmetry between image features and cluster centers, we propose an asymmetric optimal transport solver that handles source and target marginals separately.
Given a log-affinity matrix $\mathbf{M} \in \mathbb{R}^{(m+1) \times n}$ (where higher values indicate better matches), log-source marginals $\log \mathbf{a} \in \mathbb{R}^{m+1}$, and log-target marginals $\log \mathbf{b} \in \mathbb{R}^n$, our asymmetric optimal transport solver proceeds in two stages: iterative row-column normalization averaging, followed by independent marginal calibration.
\subsubsection{Initialization and Regularization} 
We initialize the log-transport matrix as:
\begin{equation}
\mathbf{Z}^{(0)} = \frac{\mathbf{M}}{\max(\tau, \epsilon)},
\end{equation}
where $\tau$ is a regularization parameter (temperature) and $\epsilon = 10^{-6}$ ensures numerical stability.
\subsubsection{Row-Column Normalization}
For each iteration $t = 1, \ldots, T$, we perform row and column normalization in the log-domain, then average the results:
\begin{equation}
\begin{aligned}
\mathbf{Z}_r^{(t)} &= \mathbf{Z}^{(t-1)} - \log\text{sumexp}(\mathbf{Z}^{(t-1)}, \text{dim}=2) \\
\mathbf{Z}_c^{(t)} &= \mathbf{Z}^{(t-1)} - \log\text{sumexp}(\mathbf{Z}^{(t-1)}, \text{dim}=1) \\
\mathbf{Z}^{(t)} &= \frac{1}{2}(\mathbf{Z}_r^{(t)} + \mathbf{Z}_c^{(t)}),
\end{aligned}
\end{equation}
where $\log\text{sumexp}(\mathbf{Z}, \text{dim}=d)$ denotes the log-sum-exp operation along dimension $d$ for numerical stability. This averaging strategy simultaneously balances row and column constraints, preventing the solution from being dominated by either dimension and enabling more stable convergence.
\subsubsection{Asymmetric Marginal Calibration}
After $T$ iterations, we perform separate marginal calibration for source and target distributions. First, we compute the source marginal calibration:
\begin{equation}
\mathbf{u} = \log \mathbf{a} - \log\text{sumexp}(\mathbf{Z}^{(T)}, \text{dim}=2),
\end{equation}
where $\log\text{sumexp}(\mathbf{Z}^{(T)}, \text{dim}=2)$ computes the log-sum of each row. We then apply the source calibration:
\begin{equation}
\mathbf{Z}' = \mathbf{Z}^{(T)} + \mathbf{u} \mathbf{1}_n^\top.
\end{equation}
Next, we compute the target marginal calibration:
\begin{equation}
\mathbf{v} = \log \mathbf{b} - \log\text{sumexp}(\mathbf{Z}', \text{dim}=1),
\end{equation}
where $\log\text{sumexp}(\mathbf{Z}', \text{dim}=1)$ computes the log-sum of each column. The final log-transport matrix is:
\begin{equation}
\log \mathbf{P} = \mathbf{Z}' + \mathbf{1}_{m+1} \mathbf{v}^\top.
\end{equation}

This formulation allows $\mathbf{u}$ and $\mathbf{v}$ to be calibrated independently, enabling asymmetric transport that adapts to different source and target distributions. Unlike standard Sinkhorn iterations that enforce symmetric marginal constraints, our approach allows the transport plan to better handle imbalanced feature distributions commonly encountered in visual place recognition, where the number of source features (clusters) may differ from the number of target features (image tokens), and their distributions may be inherently asymmetric.

\subsection{Geometric Constraints}
To leverage spatial structure in feature assignment, we introduce geometric constraints that encode spatial position information.
\subsubsection{Coordinate Embedding}
For each spatial location $(x, y)$ in the feature map, we generate normalized coordinates:
\begin{equation}
\mathbf{coord}_{xy} = \left[\frac{2x}{H-1} - 1, \frac{2y}{W-1} - 1\right] \in [-1, 1]^2
\end{equation}
These coordinates are then projected to a geometric embedding space:
\begin{equation}
\mathbf{g}_{xy} = \phi_g(\mathbf{coord}_{xy}) \in \mathbb{R}^{d_g}
\end{equation}
where $\phi_g$ is a learnable projection network (implemented as a 1×1 convolution) and $d_g$ is the geometric embedding dimension.
\subsubsection{Geometric Compatibility Score}
For each cluster center $\mathbf{c}_j$, we maintain a learnable geometric embedding $\mathbf{c}_j^g \in \mathbb{R}^{d_g}$ that represents its spatial preference. The geometric compatibility score between feature at location $(x, y)$ and cluster $j$ is:
\begin{equation}
\mathbf{S}_{ij}^g = \mathbf{g}_{xy}^\top \mathbf{c}_j^g
\end{equation}
The final score matrix is a weighted combination of feature similarity and geometric compatibility:
\begin{equation}
\mathbf{S}_{ij} = \mathbf{S}_{ij}^f + \lambda_g \mathbf{S}_{ij}^g
\end{equation}
where $\mathbf{S}_{ij}^f$ is the feature similarity score and $\lambda_g$ is a learnable scalar that controls the influence of geometric constraints.
\section{Experiments}
\subsection{Implementation Details}
We implement $A^2$GC-VPR using the PyTorch Lightning framework for modularity and reproducibility. The network employs DINOv2 ViT-B/14 \cite{zhang2022dino} as the backbone, extracting local feature maps $\mathbf{F} \in \mathbb{R}^{768 \times H \times W}$ and a global token $\mathbf{t} \in \mathbb{R}^{768}$ from $224 \times 224$ input images. We fine-tune the last 4 transformer blocks while keeping earlier blocks frozen. The aggregation module consists of three projection networks: a two-layer MLP ($\phi_t$) that projects the global token to dimension $g = 256$, a two-layer convolutional network ($\phi_c$) that projects local features to dimension, and a two-layer convolutional score network ($\phi_s$, $\psi_s$) that computes similarity scores for $m = 64$ clusters. For geometric constraints, we employ a 1×1 convolutional layer ($\phi_g$) that projects 2D coordinates to a $d_g = 16$-dimensional embedding space. Learnable parameters include geometric embeddings $\mathbf{C}^g \in \mathbb{R}^{m \times d_g}$ initialized with $\mathcal{N}(0, 0.02^2)$, a dustbin parameter $z$ initialized to 1.0, and a geometric weight $\lambda_g$ initialized to 0.15. The asymmetric optimal transport solver performs $T = 3$ iterations of row-column normalization averaging with temperature $\tau = 1.0$, followed by separate marginal calibration for numerical stability. We train on the GSV-Cities dataset \cite{ali2022gsv} containing approximately 1.2 million images from 23 cities, sampling 4 images per place. Training employs RandAugment with 3 operations, ImageNet normalization, AdamW optimizer with learning rate $6 \times 10^{-5}$ and weight decay $9.5 \times 10^{-9}$, and a linear learning rate schedule decaying to 20\% over 4000 iterations. We use MultiSimilarityLoss \cite{wang2019multi} with parameters $\alpha = 1.0$ and $\beta = 50$, combined with MultiSimilarityMiner using cosine similarity and a margin of 0.1. Training is performed with batch size 60 until convergence, with validation on Pittsburgh30k \cite{Torii-CVPR2013} validation/test sets and MSLS validation set \cite{warburg2020mapillary}. Evaluation reports Recall@K on standard VPR benchmarks, including Pittsburgh30k, MSLS, NordLand \cite{sunderhauf2013we}, and SPED. Retrieval uses FAISS \cite{douze2025faiss} with L2 distance on L2-normalized descriptors. All experiments are conducted on NVIDIA GPUs with single-GPU training (V100-32G).
\subsection{Results}
\subsubsection{Comparison with SOTA Approaches}
We compare $A^2$GC against existing SOTA VPR methods across four benchmark datasets. As shown in Table \ref{table1}, on the Pitts30k dataset, $A^2$GC achieves the best performance with Recall@1/5/10 of 95.6\%/99.3\%/99.8\%, outperforming strong competitors such as Pair-VPR (95.4\%/97.5\%/98.0\%) and CricaVPR (94.9\%/97.3\%/98.2\%). Similarly, on Pitts250k-test, $A^2$GC leads with 97.3\%/99.3\%/99.7\%, surpassing FoL and SelaVPR. On the more challenging MSLS datasets, $A^2$GC demonstrates competitive performance: on MSLS-val, it achieves 93.6\%/97.5\%/97.9\%, slightly surpassing FoL, and similar to Pair-VPR. on MSLS-challenge, it attains 80.6\%/90.9\%/92.5\%, comparable to FoL and Pair-VPR. The results validate $A^2$GC's effectiveness, particularly on urban datasets (Pitts30k, Pitts250k-test), while maintaining strong performance on diverse and challenging scenarios (MSLS). 
\begin{table*}[htbp]
\centering
\caption{Performance comparison with existing SOTA methods. \textbf{Bold} is the best result, \underline{underline} is the second-best result, and \textcolor{green}{*} indicates a two-stage re-ranking method.}
\resizebox{\linewidth}{!}{
\begin{tabular}{c||c||c||c||c||c}
\hline
 & & Pitts30k        & Pitts250k & MSLS-V       & MSLS-C \\ \cline{3-6} 
Method & Desc. size & Recall@1/5/10        & Recall@1/5/10       & Recall@1//5/10      & Recall@1/5/10       \\ \hline
NetVLAD \textcolor{blue}{CVPR'2016}\cite{arandjelovic2016netvlad} &32768 & 81.9/91.2/93.7  & 90.5/96.2/97.4 & 53.1/66.5/71.1 & 35.1/47.4/51.7 \\ 
SFRS \textcolor{blue}{ECCV'2020}\cite{ge2020self}   & 4096& 89.4/94.7/95.9  & 90.7/96.4/97.6 & 69.2/80.3/83.1 & 41.5/52.0/56.3 \\ 
PatchNetVLAD \textcolor{blue}{CVPR'2021}\cite{hausler2021patch} & 4096& 88.7/94.5/95.9  & ---            & 79.2/80.3/83.1 & 48.1/57.6/60.5 \\ 
CosPlace \textcolor{blue}{CVPR'2022} \cite{berton2022rethinking} & 2048 & 88.5/94.5/95.2  & 92.4/97.2/98.1 & 82.8/89.7/92.0 & 61.4/72.0/76.6 \\ 
MixVPR \textcolor{blue}{WACV'2023}\cite{ali2023mixvpr}&4096 & 91.5/95.5/96.3  & 94.6/98.3/99.0 & 88.2/93.1/94.3 & 64.0/75.9/80.6 \\ 
R$^2$Former \textcolor{blue}{CVPR'2023}\cite{zhu2023r2former}& 256 \textcolor{green}{*}& 91.1/ 95.2/96.3 & 93.2/97.5/98.3 & 89.7/95.0/96.2 & 73.0/85.9/88.8 \\ 
EigenPlaces \textcolor{blue}{ICCV'2023}\cite{berton2023eigenplaces}&2048 & 92.5/96.8/97.6  & 94.1/98.0/98.7 & 89.1/93.8/95.0 & 67.4/77.1/81.7 \\ 
SelaVPR \textcolor{blue}{ICLR'2024}\cite{lu2024towards}&1024 (ViTg) \textcolor{green}{*}& 92.8/96.8/97.7  & 95.7/98.8/99.2 & 90.8/96.4/97.2 & 73.5/87.5/90.6 \\ 
CricaVPR \textcolor{blue}{CVPR'2024}\cite{lu2024cricavpr} & 4096 (ViTb) \textcolor{green}{*}& 94.9/97.3/98.2  & 95.6/98.9/99.5 & 90.0/95.4/96.4 & 69.0/82.1/85.7 \\ 
TransVPR \textcolor{blue}{CVPR'2022}\cite{wang2022transvpr}&256 \textcolor{green}{*}& 89.0/94.9/96.2  &        ---        & 86.8/91.2/92.4 & 63.9/74.0/77.5 \\ 
SALAD \textcolor{blue}{CVPR'2024}\cite{izquierdo2024optimal}& 8448 (ViTb)& 92.4/96.3/97.4  & 95.1/98.5/99.1 & 92.2/96.2/97.0 & 75.0/88.8/91.3 \\ 
BoQ \textcolor{blue}{CVPR'2024}\cite{ali2024boq} & 4096& 94.5/\underline{98.9}/---     & 95.0/98.4/99.1 & 91.4/94.5/96.1 & ---            \\ 
FoL \textcolor{blue}{AAAI'2025}\cite{wang2025focus}& 8192 \textcolor{green}{*}& 94.5/97.4/98.2  & 97.0/99.2/\underline{99.5} & 93.5/96.9/97.6 & 80.0/\textbf{90.9}/\textbf{93.0} \\ 
Pair-VPR \textcolor{blue}{RAL'2025}\cite{hausler2025pair} &512 (ViTg) \textcolor{green}{*} & \underline{95.4}/97.5/98.0  & ---            & \textbf{95.4}/97.3/97.7 & \textbf{81.7}/90.2/91.3 \\ 
SemVPR \textcolor{blue}{ICCV'2025}\cite{zhang2025efficient}& 1024 (ViTl) & 94.2/97.7/98.5  & ---            & 89.7/94.7/95.9 & ---            \\ \hline 
\rowcolor{blue!10} $A^2$GC  & 2112 (ViTb) & 94.6/98.8/\underline{99.5} & 96.6/98.7/99.4& 93.0/97.3/\underline{97.8} & 79.7/90.2/92.0\\ 
\rowcolor{blue!10} $A^2$GC  & 8448 (ViTb) &  94.9/98.5/\underline{99.5} & 97.0/99.1/\textbf{99.7}& 93.3/97.4/\textbf{97.9} & 80.3/90.6/92.4 \\
\rowcolor{blue!10} $A^2$GC &33280 (ViTb) & 95.0/\underline{98.9}/ \underline{99.5}  & \underline{97.2}/\underline{99.2}/\textbf{99.7} & \underline{93.5}/\underline{97.5}/\textbf{97.9} & 80.4/\underline{90.8}/\underline{92.5} \\
\rowcolor{blue!10} $A^2$GC &33280 (ViTg) & \textbf{95.6}/\textbf{99.3}/\textbf{99.8}  & \textbf{97.3}/\textbf{99.3}/\textbf{99.7} & \underline{93.6}/\textbf{97.5}/\textbf{97.9} & \underline{80.6}/\textbf{90.9}/\underline{92.5} \\ \hline
\end{tabular}
}
\label{table1}
\end{table*}
\subsubsection{Ablation Study}
We conduct comprehensive ablation studies to validate the effectiveness of our proposed components. All experiments are performed on the Pitts30k validation set, and we report Recall@1, Recall@5, and Recall@10 metrics.

\textbf{Ablation on Different DINOV2 Backbones.} We evaluate four $A^2$GC variants with DINOV2 backbones (ViTs, ViTb, ViTl, ViTg) of 22.9M, 88.0M, 306.1M, and 1106.3M parameters on the Pitts30k dataset, as shown in Table 2. The results demonstrate a monotonically increasing trend in retrieval performance with model scale: Recall@1 improves from 94.0\% (ViTs) to 95.6\% (ViTg), Recall@5 from 98.5\% to 99.3\%, and Recall@10 from 99.3\% to 99.8\%, indicating that larger encoders learn more discriminative visual representations. However, this comes at the cost of increased computational overhead—inference latency grows from 1.32ms to 25.06ms ($19\times$ increase) while parameters increase $48\times$. Although ViTg achieves optimal performance across all metrics, ViTl (306.1M parameters, 7.85ms latency) attains near-optimal Recall@5 and Recall@10 (99.2\% and 99.7\%) with significantly lower latency, demonstrating a favorable performance-efficiency trade-off for practical deployment.
\begin{table}[htbp]
\centering
\caption{DINOv2 configurations and performances.}
\small
\begin{tabular}{cccccc}
\hline
                                  &                             &                           & \multicolumn{3}{c}{Pitts30k}                                             \\ \cline{4-6} 
Method & {Params.} & {Latency} & \multicolumn{1}{c}{R@1} & \multicolumn{1}{c}{R@5} & R@10 \\ \hline
ViTs                                      & 22.9   M                    & 1.32   ms                 & \multicolumn{1}{c}{94.0}     & \multicolumn{1}{c}{98.5}     & 99.3      \\ 
ViTb                                      & 88.0   M                    & 2.41   ms                 & \multicolumn{1}{c}{94.9}     & \multicolumn{1}{c}{98.5}     & 99.5      \\ 
ViTl                                     & 306.1   M                   & 7.85   ms                 & \multicolumn{1}{c}{95.4}     & \multicolumn{1}{c}{99.2}     & 99.7      \\ 
ViTg                                      & 1106.3   M                  & 25.06   ms                & \multicolumn{1}{c}{\textbf{95.6}}     & \multicolumn{1}{c}{\textbf{99.3}}     & \textbf{99.8}      \\ \hline
\end{tabular}
\end{table}

\textbf{Ablation Study on Descriptor Size.} As shown in Table 3, we evaluate four $A^2$GC configurations with varying descriptor sizes (512+32, 2048+64, 8192+256, 32768+512) on the Pitts30k dataset. The results reveal that retrieval performance generally improves with descriptor size: Recall@1 increases from 93.7\% to 95.0\%, indicating that larger descriptors capture richer discriminative information for top-1 retrieval accuracy. Recall@5 shows an overall upward trend from 98.7\% to 98.9\%, with a slight dip at the (8192+256) configuration (98.5\%), suggesting potential feature redundancy at intermediate sizes. Recall@10 quickly saturates at 99.5\% after the (2048+64) configuration and remains stable for larger descriptors, demonstrating that high-rank recall benefits less from increased descriptor dimensionality. The largest configuration (32768+512) achieves optimal performance on Recall@1 (95.0\%) and Recall@5 (98.9\%), while matching the best Recall@10 (99.5\%), validating the effectiveness of larger descriptors for high-precision visual localization tasks.
\begin{table}[htbp]
\centering
\caption{Performance evaluation for different descriptor sizes with the training of DINOV2 vitb backbone.}
\begin{tabular}{cccc}
\hline
& \multicolumn{3}{c}{Pitts30k}                                             \\ \cline{2-4} 
Method (Descriptor Size) & \multicolumn{1}{c}{R@1} & \multicolumn{1}{c}{R@5} & R@10 \\ \hline
512+32                                                           & \multicolumn{1}{c}{93.7}     & \multicolumn{1}{c}{98.7}     & 99.4      \\ 
2048+64                                                          & \multicolumn{1}{c}{94.6}     & \multicolumn{1}{c}{98.8}     & \textbf{99.5}      \\ 
8192+256                                                         & \multicolumn{1}{c}{94.9}     & \multicolumn{1}{c}{98.5}     & \textbf{99.5}      \\ 
32768+512                                                        & \multicolumn{1}{c}{\textbf{95.0}}     & \multicolumn{1}{c}{\textbf{98.9}}     & \textbf{99.5}      \\ \hline
\end{tabular}
\end{table}

\textbf{Ablation Study on Different Fine-tuning Strategies.} We evaluate five $A^2$GC configurations with different fine-tuning strategies (frozen, train 2/4/6 last blocks, train all blocks) on the Pitts30k dataset. As shown in Table 4, results demonstrate that fine-tuning significantly improves performance over the frozen baseline: Recall@1 increases from 92.9\% (frozen) to 94.9\% (train 4 last blocks), Recall@5 from 98.3\% to 99.0\% (train 2/6 last blocks), and Recall@10 from 99.0\% to 99.6\% (train 2/all blocks). Notably, training only the last 2-4 blocks achieves optimal or near-optimal performance across all metrics, with training the last 4 blocks yielding the best Recall@1 (94.9\%) and training the last 2 blocks achieving the best Recall@5 (99.0\%) and Recall@10 (99.6\%). In contrast, training all blocks results in slightly degraded performance (Recall@1: 94.0\%), suggesting that excessive fine-tuning may lead to overfitting or disrupt the pre-trained feature representations. 
\begin{table}[htbp]
\centering
\caption{Performance evaluation with different fine-tuning strategies with the training of DINOV2 vitb backbone.}
\begin{tabular}{cccc}
\hline
 & \multicolumn{3}{c}{Pitts30k}                                             \\ \cline{2-4} 
Method        & \multicolumn{1}{c}{R@1} & \multicolumn{1}{c}{R@5} & R@10 \\ \hline
 frozen                  & \multicolumn{1}{c}{92.9}     & \multicolumn{1}{c}{98.3}     & 99.0      \\
train 2 last blocks    & \multicolumn{1}{c}{94.8}     & \multicolumn{1}{c}{\textbf{99.0}}     & 99.6      \\ 
train   4 last blocks & \multicolumn{1}{c}{\textbf{94.9}}     & \multicolumn{1}{c}{98.5}     & 99.5      \\ 
train 6 last blocks     & \multicolumn{1}{c}{94.6}     & \multicolumn{1}{c}{\textbf{99.0}}     & 99.5      \\ 
train all blocks     & \multicolumn{1}{c}{94.0}     & \multicolumn{1}{c}{98.9}     & \textbf{99.6}      \\ \hline
\end{tabular}
\end{table}
\begin{figure*}
    \centering
    \includegraphics[width=0.95\linewidth]{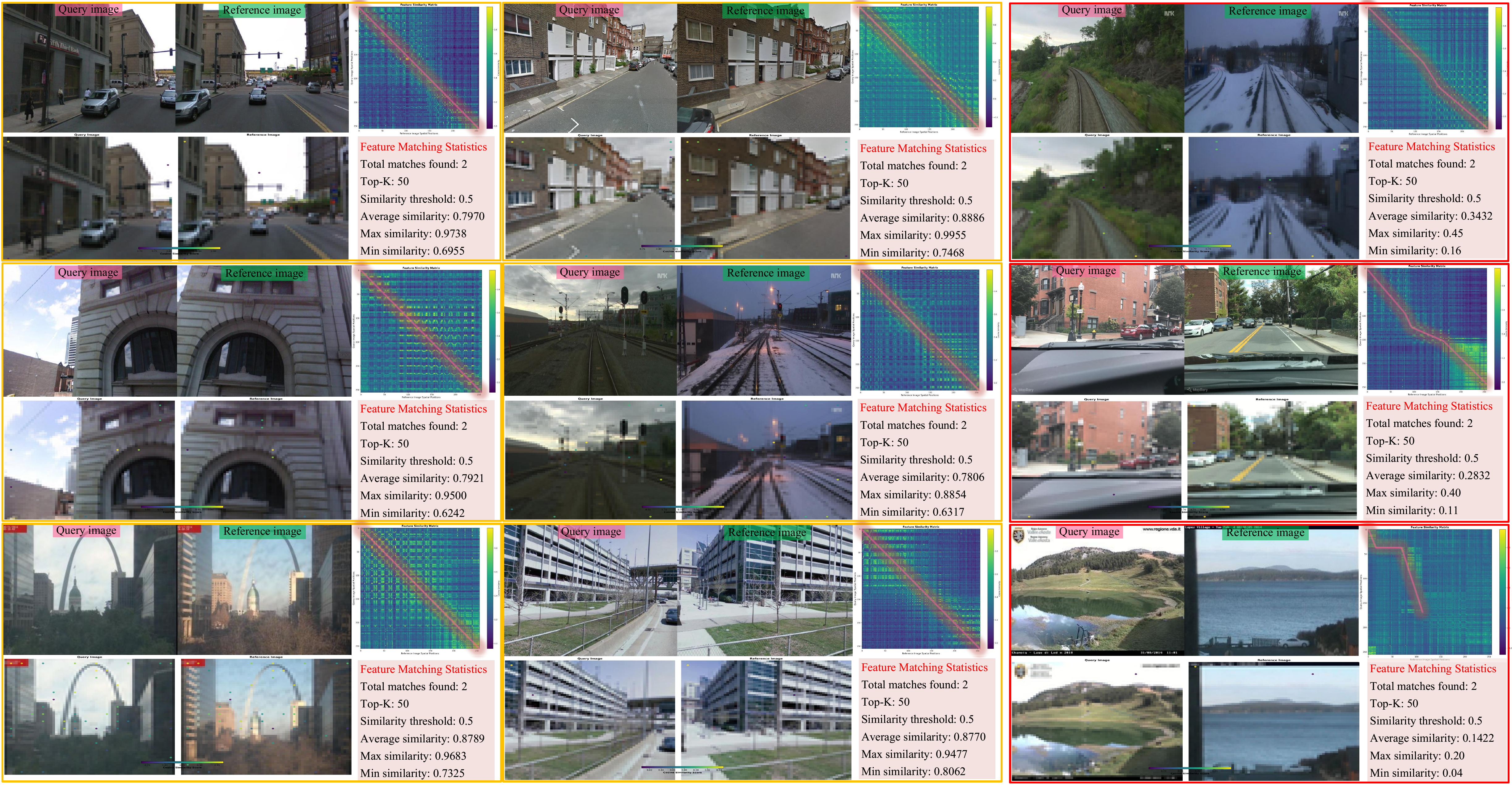}
    \caption{Visualization analysis of feature activation and image matching. We masked activate regions using DINOV2 encodering and asymmetric aggregation. We also utilized cosine similarity to achieve confusion (similarity) matrix. \textcolor{yellow}{yellow boxes} denote correctly matched query–reference pairs and \textcolor{red}{red boxes} indicate failures.}
\end{figure*}

\textbf{Ablation Study on Component Contributions.} To analyze the contribution of each component in $A^2$GC, we conduct ablation studies on the Pitts30k dataset by removing Asymmetric Aggregation ($A^2$GC) and Geometric Constraints (GC) components, as shown in Table 5. The full $A^2$GC method achieves the best performance with Recall@1/5/10 of 94.9\%/98.5\%/99.5\%, demonstrating the effectiveness of both components. Removing both $A^2$GC and GC results in the lowest performance (92.5\%/96.4\%/97.8\%), indicating that both components are essential for optimal performance. When removing $A^2$GC alone, performance drops to 93.9\%/98.1\%/99.3\%, while removing GC alone yields 94.1\%/97.9\%/99.5\%. The results show that $A^2$GC has a slightly larger impact on Recall@1 (1.0\% drop vs. 0.8\% drop when removing GC), while GC contributes more to Recall@5 performance (0.6\% drop vs. 0.4\% drop when removing $A^2$GC). These findings validate that both Asymmetric Aggregation and Geometric Constraints play complementary roles in enhancing visual place recognition performance.
\begin{table}[htbp]
\centering
\caption{Performance evaluation for key components of the proposed $A^2$GC with the training of DINOV2 vitb backbone.}
\begin{tabular}{cccc}
\hline
  & \multicolumn{3}{c}{Pitts30k}                                             \\ \cline{2-4} 
Method & \multicolumn{1}{c}{R@1} & \multicolumn{1}{c}{R@5} & R@10 \\ \hline
w/o $A^2$                                           & \multicolumn{1}{c}{93.9}     & \multicolumn{1}{c}{98.1}     & 99.3      \\ 
w/o GC                                          & \multicolumn{1}{c}{94.1}     & \multicolumn{1}{c}{97.9}     & \textbf{99.5}      \\ 
w/o $A^2$ \& GC                                     & \multicolumn{1}{c}{92.5}     & \multicolumn{1}{c}{96.4}     & 97.8      \\ 
$A^2$GC                                                      & \multicolumn{1}{c}{\textbf{94.9}}     & \multicolumn{1}{c}{\textbf{98.5}}     & \textbf{99.5}      \\ \hline
\end{tabular}

\end{table}

\textbf{Ablation Study on Image Size.} Table 6 demonstrates a strong positive correlation between input resolution and VPR performance across both datasets. As resolution increases from $112\times112$ to $588\times588$ pixels, recall metrics exhibit consistent improvements, with the most substantial gains observed on the more challenging MSLS-val dataset (Recall@1 improves from 81.6\% to 96.0\%, a +14.4\% increase) compared to Pitts30k (from 92.1\% to 96.7\%, a +4.6\% increase). Higher resolutions preserve fine-grained visual details and multi-scale information crucial for distinguishing visually similar locations, particularly in complex urban environments. However, performance gains exhibit diminishing returns at very high resolutions, suggesting a trade-off between accuracy and computational efficiency. 
\begin{table}
\centering
\caption{Experimental evaluation for different input image sizes with the DINOV2\_ViTb backbone.}
\begin{tabular}{ccc}
\hline
{Input   size} & Pitts30k       & MSLS-val       \\ \cline{2-3} 
                              & R@1/5/10       & R@1/5/10       \\ \hline
112                           & 92.1/97.9/98.8 & 81.6/90.3/92.4 \\ 
224                           & 94.9/98.5/99.5 & 90.4/95.3/96.1 \\ 
364                           & 94.9/99.1/99.6 & 91.0/96.0/96.6 \\
406                           & 95.2/99.2/99.8 & 93.2/96.7/97.2 \\ 
588                           & \textbf{96.7/99.8/100}  & \textbf{96.0/97.9/98.6} \\ \hline
\end{tabular}

\end{table}

\subsection{Interesting findings}
The method exhibits a "human-like" attention mechanism, naturally ignoring irrelevant areas like the sky and focusing on stable features crucial for VPR, such as road surfaces and structures, as shown in Figure 3. Activation patterns reveal both global scene understanding and local feature discrimination. Notably, the feature correlation between query and reference images remains consistent across different viewpoints, even with dynamic objects like vehicles. This suggests the model extracts viewpoint-invariant scene signatures, capturing geometric coherence. The consistent activations in stable regions across images demonstrate the model’s ability to match scene features robustly.
\subsection{Test Experiments}
We evaluate generalization performance on two challenging test sets: Nordland (spring vs. winter) with extreme seasonal variations, and SPED with diverse urban scenes. As shown in Table 7, $A^2$GC consistently outperforms all competitors across both datasets. On Nordland, $A^2$GC achieves R@1/5/10 of 34.71\%/49.71\%/56.20\%, substantially outperforming SALAD and dramatically exceeding other methods, achieving approximately $1.4\times$ higher Recall@1 than SALAD. On SPED, $A^2$GC maintains leading performance with 83.86\%/92.75\%/94.56\%, outperforming SALAD and significantly surpassing other methods. The consistent superiority validates the effectiveness of asymmetric aggregation and geometric constraints in handling domain shifts and appearance variations. \\
\indent Also, as shown in Table 8, we test some SOTA-trained models on the SF-XL dataset. The proposed $A^2$GC still achieves unprecedented results on the SF-XL dataset despite never seeing any single image during VPR finetuning.\\
\indent Finally, in Figure 4, we present various query images along with the top-1 retrieval result for each query image as predicted by existing SOTA methods. $A^2$GC demonstrates the ability to retrieve accurate results even under challenging conditions, such as significant variations in lighting or viewpoint.
\begin{table}[htbp]
\centering
\caption{Test experiments. To further validate the performance and robustness of the proposed method in challenging scenarios, we used models trained on the GSV-Cities dataset (including MixVPR, FoL, CricaVPR, SALAD, and Pair-VPR) for Nordrdland (\textcolor{red}{spring} \textcolor{gray}{VS} \textcolor{blue}{winter}) and SPED tests. \textit{The proposed $A^2$GC achieves unprecedented results on Nordrdland and SPED despite never seeing any single image during VPR finetuning.}}
\small
\begin{tabular}{cccc}
\hline
Method & Nordland (spr. vs win.) & SPED              \\ \cline{2-3} 
                        & R@1/5/10                      & R@1/5/10          \\ \hline
MixVPR                  & 4.82/8.01/8.01                & 63.59/80.56/84.51 \\
FoL                     & 6.20/11.70/14.60              & 67.38/83.20/86.33 \\ 
CricaVPR                & 6.12/11.23/13.84                & 64.25/64.25/85.34 \\ 
SALAD                   & 25.18/37.43/42.07             & 81.71/91.76/95.06 \\ 
Pair-VPR                & 8.12/14.67/17.86                & 65.24/79.90/83.69 \\ 
$A^2$GC                 & \textbf{34.71}/\textbf{49.71}/\textbf{56.20}      & \textbf{83.86}/\textbf{92.75}/\textbf{95.56}  \\ \hline
\end{tabular}

\end{table}
\begin{table}
\centering
\caption{Top-1 Recall test on SF-XL dataset. During the testing process, no images were ever seen during VPR fine-tuning.}
\small
\begin{tabular}{cccc}
\hline
Method      & Size & SF-XL   (v1\_test) & SF-XL   (v2\_test) \\ \hline
CosPlace    & 2048              & 76.4               & 88.8               \\ 
EigenPlaces & 2048              & 84.1               & 90.8               \\ 
SALAD       & 8448              & 88.6               & 94.8               \\ 
$A^2$GC        & 8448              & \textbf{89.5}               & \textbf{95.2}               \\ \hline
\end{tabular}

\end{table}
\begin{figure}
    \centering
    \includegraphics[width=1.0\linewidth]{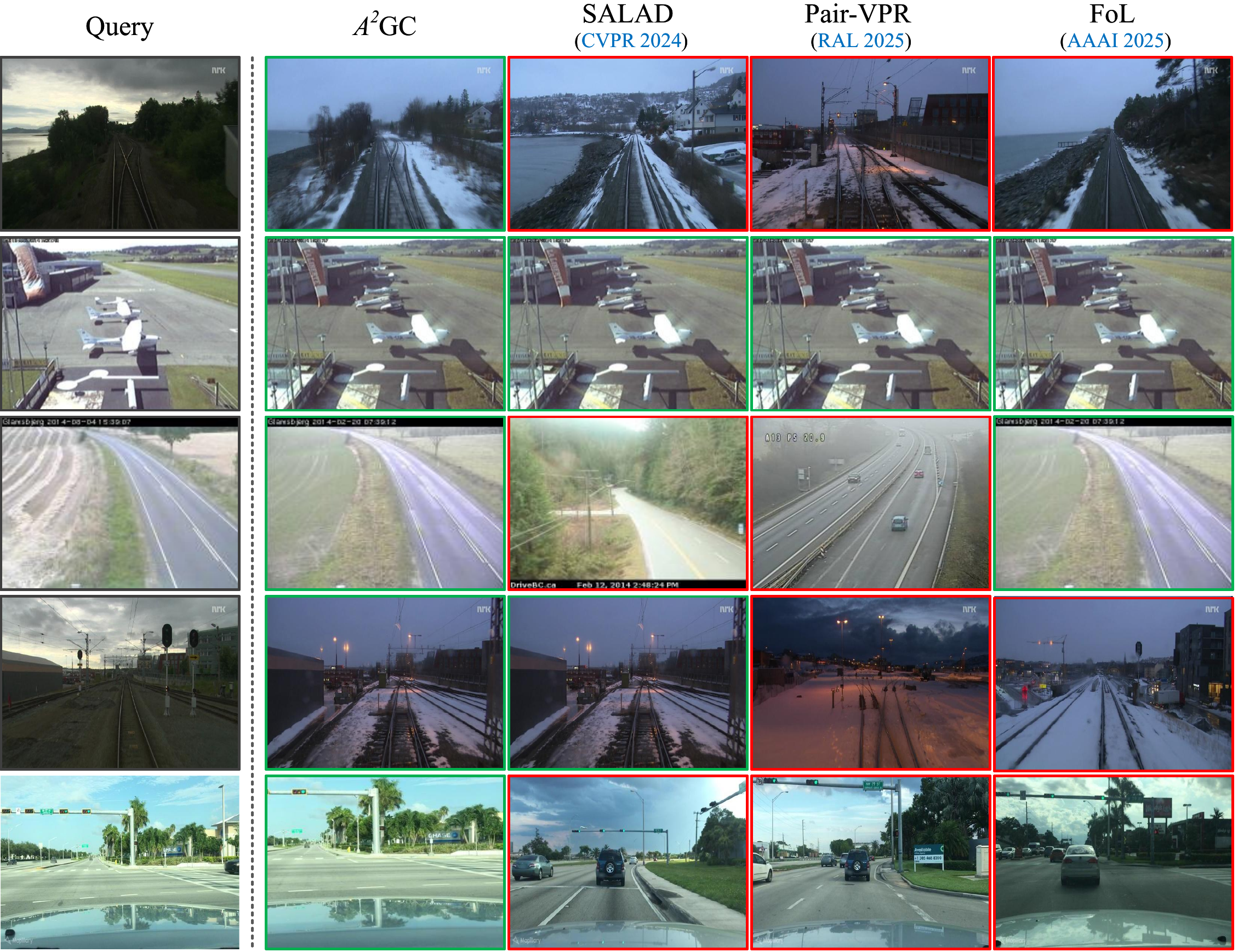}
    \caption{Qualitative results at challenging datasets. The left column shows several queries, and the right columns show the top-1 candidate retrieved by existing SOTA methods.}

\end{figure}
\section{Conclusions}
We present $A^2$GC, a VPR method combining asymmetric optimal transport and geometric constraints. The asymmetric transport solver adaptively handles imbalanced feature distributions via independent marginal calibration, while geometric constraints enforce spatial coherence through compatibility modeling between 2D coordinates and learnable cluster centers. $A^2$GC achieves SOTA results on urban datasets and demonstrates exceptional cross-domain generalization, achieving approximately $1.4\times$ higher Recall@1 than existing SOTA methods on Nordland with extreme seasonal or illumination variations. Ablation studies confirm both components are essential and complementary: asymmetric aggregation improves Recall@1 by handling distribution mismatches, while geometric constraints enhance mid-rank accuracy. 

{
\bibliographystyle{ieeenat_fullname}

}


\end{document}